\documentclass[10pt,twocolumn,letterpaper]{article}

\usepackage[pagenumbers]{cvpr}
\makeatletter
\patchcmd{\@maketitle}{\vskip .375in}{\vskip .05in}{}{}
\patchcmd{\@maketitle}{\vspace*{24pt}}{\vspace*{8pt}}{}{}
\patchcmd{\@maketitle}{\vskip .5em}{}{}{}
\makeatother

\usepackage{graphicx}
\usepackage{amsmath,amssymb,booktabs,multirow,xspace}
\usepackage{xcolor}
\usepackage{colortbl}
\usepackage{capt-of}
\usepackage{float}   %
\definecolor{rowlit}{gray}{0.92}
\colorlet{rowteacher}{blue!8}
\colorlet{rowdistill}{orange!14}
\usepackage{microtype}

\newcommand{\mode}[1]{\underline{\textsc{#1}}\xspace}
\newcommand{\stereogs}{\mode{StereoGS}}
\newcommand{\dropg}{\mode{DropGaussian}}
\newcommand{\fsgs}{\mode{FSGS}}
\newcommand{\depthsplat}{\mode{DepthSplat}}
\newcommand{\anysplat}{\mode{AnySplat}}
\newcommand{\difix}{\mode{Difix3D+}}
\newcommand{\lager}{\mode{LagerNVS}}
\newcommand{\seva}{\mode{SEVA}}
\newcommand{\lvsm}{\mode{LVSM}}
\newcommand{\dngaussian}{\mode{DNGaussian}}
\newcommand{\binocular}{\mode{Binocular3DGS}}
\newcommand{\corgs}{\mode{CoR-GS}}
\newcommand{\dropoutgs}{\mode{DropoutGS}}
\newcommand{\instantsplat}{\mode{InstantSplat}}
\newcommand{\pixelsplat}{\mode{pixelSplat}}
\newcommand{\mvsplat}{\mode{MVSplat}}
\newcommand{\gslrm}{\mode{GS-LRM}}
\newcommand{\reconfusion}{\mode{ReconFusion}}
\newcommand{\gsgs}{\mode{GS-GS}}
\newcommand{\enhancer}{\mode{3DGS-Enhancer}}
\newcommand{\genfusion}{\mode{GenFusion}}
\newcommand{\gaps}{\mode{GAPS}}
\newcommand{\freenerf}{\mode{FreeNeRF}}
\newcommand{\fregs}{\mode{FreGS}}
\newcommand{\dwtgs}{\mode{DWTGS}}
\newcommand{\teacher}{\mode{Ours-Teacher}}
\newcommand{\distilled}{\mode{Ours-Distilled}}
\usepackage{etoolbox}
\AtBeginEnvironment{tabular}{\renewcommand{\mode}[1]{#1\xspace}\renewcommand{\teacher}{Teacher\xspace}\renewcommand{\distilled}{Distilled\xspace}}
\DeclareMathOperator*{\argmin}{arg\,min}
\newcommand{\Mean}{\mu}
\newcommand{\Fig}[1]{Fig.~\ref{fig:#1}}
\newcommand{\Tbl}[1]{Tab.~\ref{tab:#1}}
\newcommand{\Eqn}[1]{Eq.~(\ref{eq:#1})}
\newcommand{\Sect}[1]{Sec.~\ref{sec:#1}}
\newcommand{\Supp}[1]{Supp.~\ref{sec:supp:#1}}

\definecolor{cvprblue}{rgb}{0.21,0.49,0.74}

\usepackage[pagebackref,breaklinks,colorlinks,allcolors=cvprblue]{hyperref}

\title{Enhancing Sparse-View Gaussian Splatting with Feed-Forward Priors via Spectral Fusion and Distillation}

\author{Weikai Lin\qquad Jun Dai\qquad Sheng Zhao\qquad Junjie Zhao\qquad Yuhao Zhu\\
University of Rochester\\
{\tt\small wlin33@ur.rochester.edu}
}

\begin{document}
\twocolumn[{%
\renewcommand\twocolumn[1][]{#1}%
\maketitle
\begin{center}
  \vspace{-1.5em}
  \includegraphics[width=\linewidth]{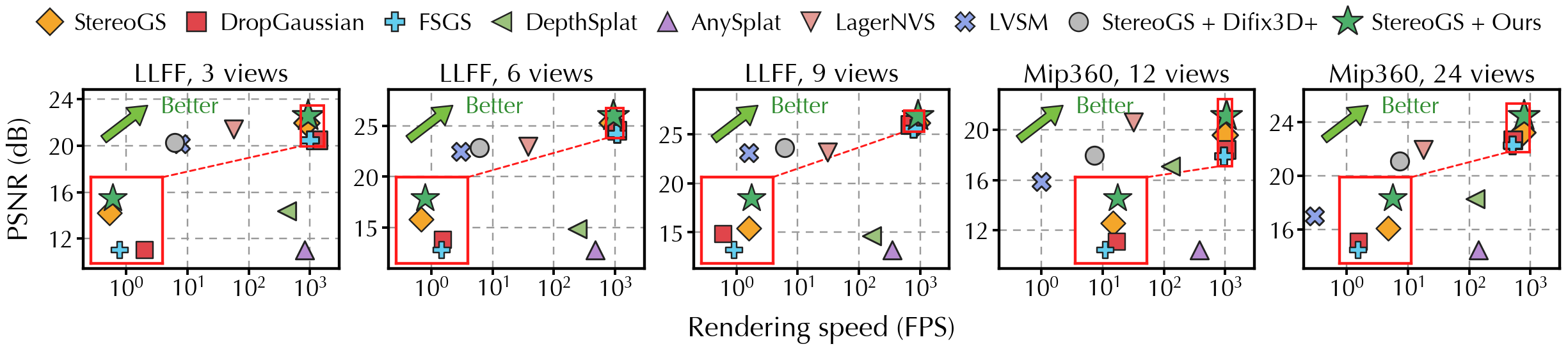}
  \captionof{figure}{Our method enhances existing sparse-view Gaussian models with a feed-forward prior.
  Each panel plots the quality against the rendering speed on an RTX 5090 for one dataset and view count.
  Our method reaches the highest quality at more than 900 FPS.}
  \label{fig:teaser}
\end{center}%
}]
\begin{abstract}
Gaussian splatting from sparse training views renders high-quality images near training views, but reconstruction quality degrades at distant regions, which are undersampled by the training set.
Feed-forward novel view synthesis (NVS) models use strong learned priors to resample the scene from substantially different viewpoints, but they take more than ten times as long to render and produce high-frequency artifacts on out-of-domain scenes.
We propose a plug-and-play method that enhances an existing sparse-view Gaussian model with a pretrained feed-forward NVS model.
We observe that the two are complementary in the frequency domain: the prior is more accurate at low frequencies, and the Gaussians at high frequencies.
The reliability of the prior relative to the Gaussians is inversely proportional to the image frequency, so the optimal fusion weight is a first-order low-pass filter with one parameter that can be fitted.
Our key idea is then to intelligently fuse the reliable part of each of the two methods.
To remove the slow feed-forward model at render time, we use these fused images as teachers and distill them back into Gaussians, balanced with supervision from input views.
Our method improves the quality of various sparse-view Gaussian methods without sacrificing rendering speed, and our analysis works on different prior models.
Project page: \href{https://linwk20.github.io/spectral-fusion-gs/}{linwk20.github.io/spectral-fusion-gs}
\end{abstract}

\section{Introduction}
\label{sec:intro}

3D Gaussian Splatting (3DGS)~\cite{kerbl20233d} renders photorealistic novel views in real time~\cite{zhu2026seele} and is used in XR~\cite{lin2025metasapiens,lin2025powergs,lin2026controlgs,franke2025vrsplatting,tu2025vrsplat}, streaming~\cite{sun20243dgstream,yan2025instant,hu20254dgc,wu2025immersive}, autonomous driving~\cite{yan2024street,hess2025splatad,chen2025omnire} and teleoperation~\cite{wildersmith2024radiance}.
However, with only a few input views, it overfits to the inputs and performs poorly at novel views.
Sparse-view methods regularize the optimization with dropout~\cite{park2025dropgaussian,xu2025dropoutgs}, depth~\cite{zhu2024fsgs,li2024dngaussian} or stereo priors~\cite{han2024binocular,yuan2026stereogs}.
They stay sharp near the training views and break at views far from them.
Feed-forward novel view synthesis (NVS) models~\cite{jin2025lvsm,szymanowicz2026lagernvs} carry a prior learned from many scenes.
However, they cannot replace Gaussians, because (1) they render more than $10\times$ slower, and (2) their results carry high-frequency artifacts on out-of-domain scenes {(\Sect{mot:artifact})}.

This paper targets Gaussian splatting with sparse inputs.
We introduce a plug-and-play method that enhances any sparse-view Gaussian model with a pretrained feed-forward NVS model (NVS prior).
Our key idea is to combine the strengths of the two methods.
It combines three ingredients.

\paragraph{Spectral analysis.}
We compare the errors of the two models at every spatial frequency (\Sect{motivation}).
Both methods have high errors at high frequencies.
We find that the error of the Gaussians rises on a straight line in log frequency, while the error of the NVS prior starts lower and rises faster.
The two curves cross once: before the crossover, the prior has the lower error, and after it, the Gaussians are more reliable.

\paragraph{Spectral fusion.}
We derive the optimal way to combine them in \Sect{kernel}.
Given a real image, a render by the Gaussians and a render by the prior, we can compute the optimal weighting across frequency.
We find that the relative reliability of the NVS prior falls with frequency.

However, the measured optimal weight is noisy.
To characterize the optimum in a principled way and to improve generalization, we parameterize the optimal weight with a curve.
Our key finding is that the optimal fusion weight is well approximated by a first-order low-pass filter, which has only a single parameter: the crossover frequency where both models are equally reliable.
We fit this single-parameter curve on a development set and find that it generalizes well to the held-out test set and provides high-quality fused results.

\begin{figure}[t]
  \centering
  \includegraphics[width=\linewidth]{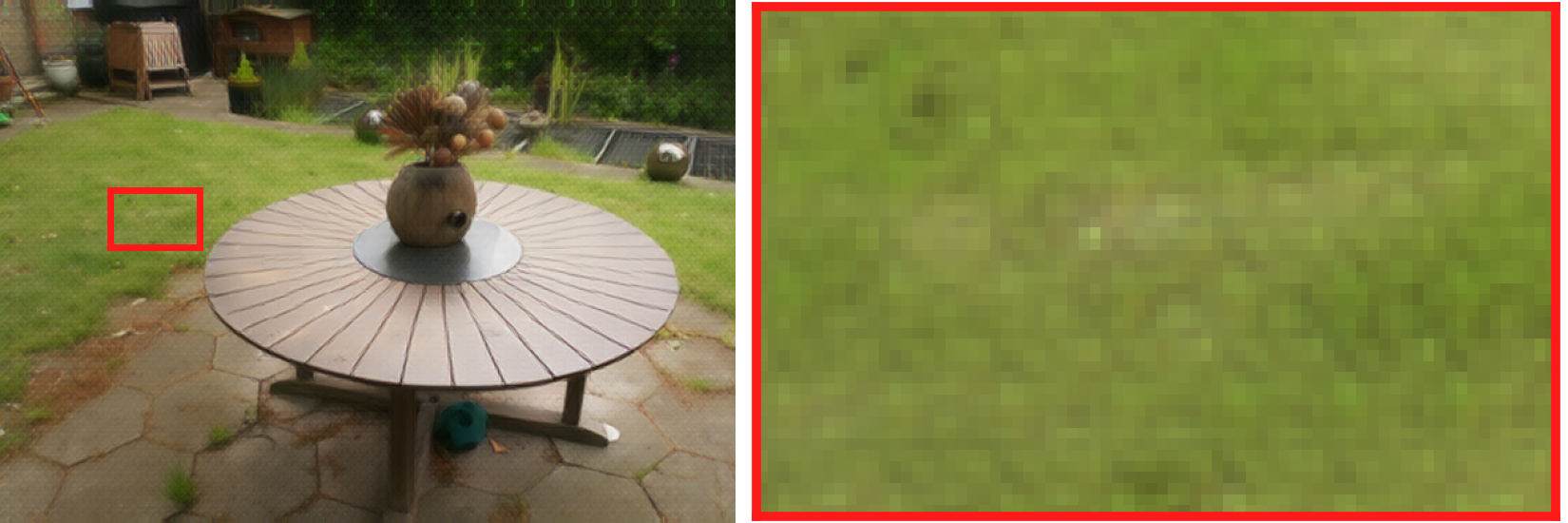}
  \caption{\lager renders high-frequency artifacts on the \texttt{garden} scene with 12 views{ (red box, enlarged on the right)}.}
  \label{fig:artifact}
\end{figure}

\paragraph{Distillation.}
Obtaining the fused results requires running the feed-forward model, which is slow.
To keep rendering as fast as Gaussians, we use the fused images as teachers (\Sect{teacher}), mix them with the real training views, and distill {this hybrid supervision into the Gaussians} (\Sect{distill}).

\paragraph{Result.}
We evaluate on LLFF and Mip-NeRF 360 with different input views.
The distilled models improve in PSNR, SSIM and LPIPS (\Fig{teaser}) by $1.0$ to $1.2$\,dB on average and render as fast as their base models, at $917$\,FPS on average for \stereogs on an RTX 5090.
Our contributions are:
\begin{itemize}
  \item We show that a feed-forward NVS prior and sparse-view Gaussians are complementary in frequency.
  \item We characterize and fit the reliability of the prior relative to the Gaussians with a parameterized first-order low-pass filter, which is used to fuse the results of the two methods.
  \item We propose a plug-and-play distillation that distills the fused results back into the Gaussians, enabling both high-quality and high-efficiency rendering.
  \item Our method improves different Gaussian methods without sacrificing rendering speed. It also generalizes to different methods that provide an NVS prior.
\end{itemize}

\section{Related Work}
\label{sec:related}

\begin{figure}[t]
  \centering
  \includegraphics[width=\linewidth]{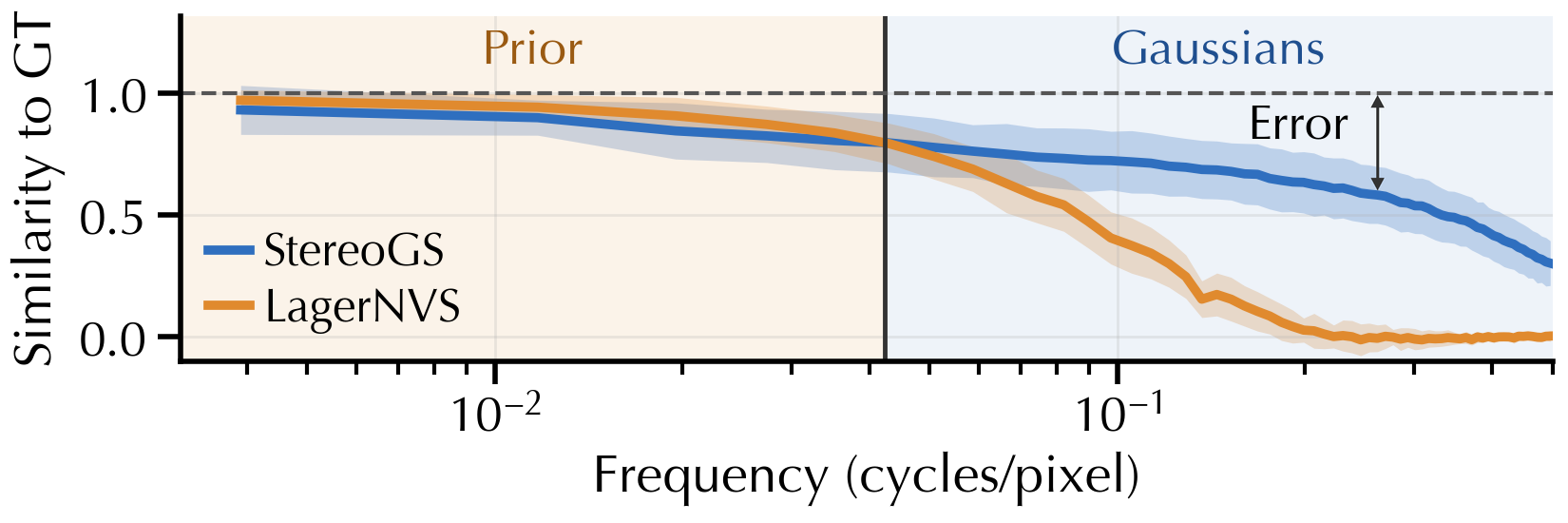}
  \caption{Similarity of two methods to the ground truth per frequency on \texttt{garden} with 12 views.}
  \label{fig:similarity}
\end{figure}
\subsection{Novel View Synthesis (NVS) from Sparse Views}

NeRF~\cite{mildenhall2020nerf} and 3D Gaussian Splatting~\cite{kerbl20233d} optimize one scene at a time and need dense views.
NeRF-based methods regularize the radiance field under sparse inputs~\cite{jain2021dietnerf,niemeyer2022regnerf,yang2023freenerf,wang2023sparsenerf}.
Sparse-view 3DGS follows two directions.
One regularizes the optimization: \dropg~\cite{park2025dropgaussian} and \dropoutgs~\cite{xu2025dropoutgs} drop Gaussians during training, and \corgs~\cite{zhang2024corgs} penalizes disagreement between two models.
The other adds geometric priors: \fsgs~\cite{zhu2024fsgs} and \dngaussian~\cite{li2024dngaussian} use monocular depth, \binocular~\cite{han2024binocular} builds stereo pairs from the inputs, \stereogs~\cite{yuan2026stereogs} uses stereo and multi-view stereo priors, and \instantsplat~\cite{fan2024instantsplat} starts from feed-forward geometry.
These methods add no appearance information at views the inputs do not cover.
Our method takes their trained model as input and leaves their training unchanged.
\subsection{Novel View Synthesis with Prior}

Feed-forward models learn to synthesize views across scenes, which removes the need for dense views.
\pixelsplat~\cite{charatan2024pixelsplat}, \mvsplat~\cite{chen2024mvsplat}, \gslrm~\cite{zhang2024gslrm}, \depthsplat~\cite{xu2025depthsplat} and \anysplat~\cite{jiang2025anysplat} predict Gaussians from images.
Feed-forward NVS~{\cite{jin2025lvsm,wang2025clift,lin2026glodlvsm}}, such as \lager~\cite{szymanowicz2026lagernvs}, predict target images directly with no explicit 3D representation.
They render slowly, so we use one only as a frozen teacher.

Diffusion models generate content at novel views, and \seva~\cite{zhou2025seva} synthesizes them directly with a multi-view diffusion model; we test it as a prior in \Sect{prior}.
\mode{ControlHair}~\cite{lin2026controlhair} guides a diffusion model with control signals and renders bullet-time novel views of hair.
\reconfusion~\cite{wu2024reconfusion} and \gsgs~\cite{kong2025generative} regularize the reconstruction with a diffusion prior.
\enhancer~\cite{liu20243dgsenhancer}, \difix~\cite{wu2025difix3d} and \genfusion~\cite{wu2025genfusion} restore artifact-prone renders and feed them back into reconstruction.
\gaps~\cite{zhu2026gaps} selects pseudo views by informativeness and generative reliability.
These methods use the prior at all frequencies.
We use a deterministic feed-forward model and weight it by frequency, because it is reliable only at low frequencies.

\subsection{Frequency Analysis in View Synthesis}

Frequency is a common tool to regularize view synthesis.
\freenerf~\cite{yang2023freenerf} masks the high-frequency positional encoding early in training, and \fregs~\cite{zhang2024fregs} and \dwtgs~\cite{nguyen2025dwtgs} supervise Gaussians progressively or separately across frequency bands.
These methods treat frequency as a schedule for one model.
In contrast, we use frequency to decide which of two models to trust.
This is the setting of classical linear estimation: the optimal weight between two noisy estimates at each frequency follows their relative reliability~\cite{wiener1949extrapolation}.
We measure this weight for a feed-forward prior and sparse-view Gaussians, and find that a one-parameter filter describes it.

\begin{figure}[t]
  \centering
  \includegraphics[width=\linewidth]{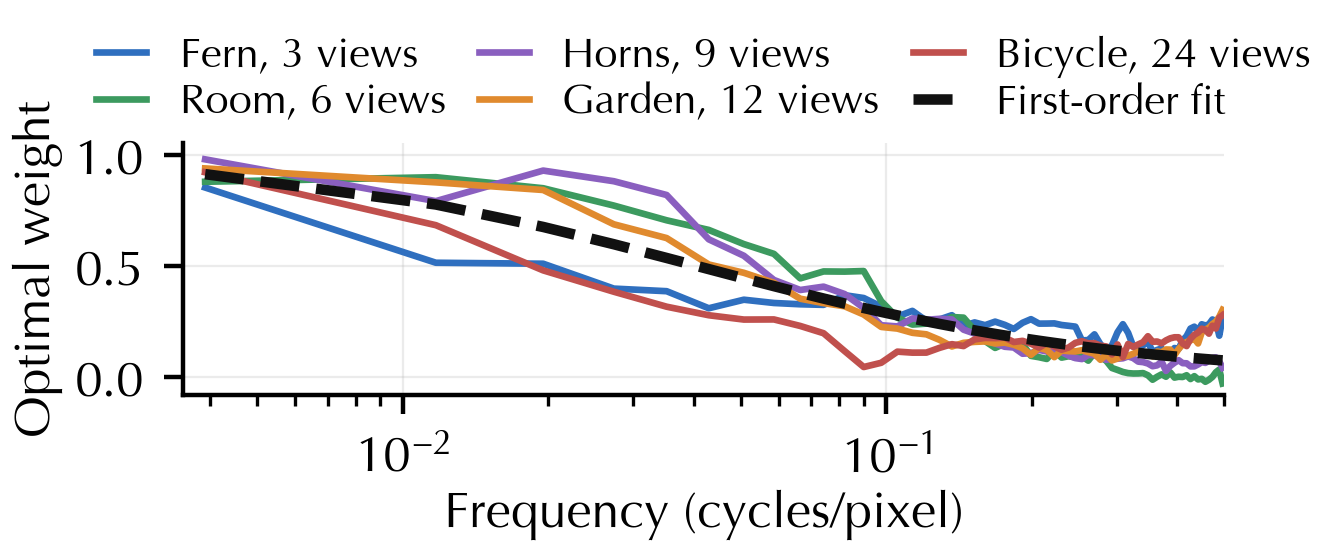}
  \caption{Optimal weight on the prior per frequency, for \stereogs on the five development settings.}
  \label{fig:optw}
  \vspace{-10pt}
\end{figure}

\section{Motivation}
\label{sec:motivation}

\paragraph{Feed-forward priors fail as direct supervision.}
\label{sec:mot:artifact}
A straightforward way to help sparse-view Gaussians is to supervise them with the outputs of a model that carries a prior~\cite{wu2024reconfusion,kong2025generative,liu20243dgsenhancer,wu2025difix3d,wu2025genfusion}.
However, we find that this does not work well with a feed-forward NVS model.
This is because on scenes that differ from its training set, the quality of a feed-forward NVS model can fall below that of a state-of-the-art sparse-view Gaussian method.
For example, we compare the state-of-the-art feed-forward \lager~\cite{szymanowicz2026lagernvs} with the sparse-view Gaussian \stereogs~\cite{yuan2026stereogs} on the Mip-NeRF 360 \texttt{garden} scene with 12 input views: averaged over the test views, \stereogs reaches $21.38$\,dB, while \lager reaches only $21.20$\,dB.
{One reason is that \lager produces high-frequency artifacts (\Fig{artifact}).}

\paragraph{Prior and Gaussians are complementary in frequency.}
\label{sec:mot:pattern}
We further measure, on this scene, how similar the renders of \lager and of \stereogs are to the ground truth at every spatial frequency (\Fig{similarity}).
The similarity is the \textit{correlation between the Fourier coefficients} of a render and of the ground truth within each frequency ring, and we plot its mean and standard deviation over the test views.
As the frequency increases, the similarity between \lager and the ground truth drops rapidly.
The similarity of \stereogs also decreases with frequency, but gradually.
The two curves cross once.
The same pattern holds over different settings and Gaussian methods (\Sect{analysis}).

\label{sec:mot:implication}
These results define a frequency threshold: below it, \lager is more reliable than the Gaussians, and above it, the Gaussians are more reliable than \lager (shaded regions in \Fig{similarity}).
The prior of a feed-forward NVS model is therefore not useless: it must be used in the specific frequency band where it is reliable.

\section{Method}
\label{sec:method}

\begin{figure*}[t]
  \centering
  \includegraphics[width=\linewidth]{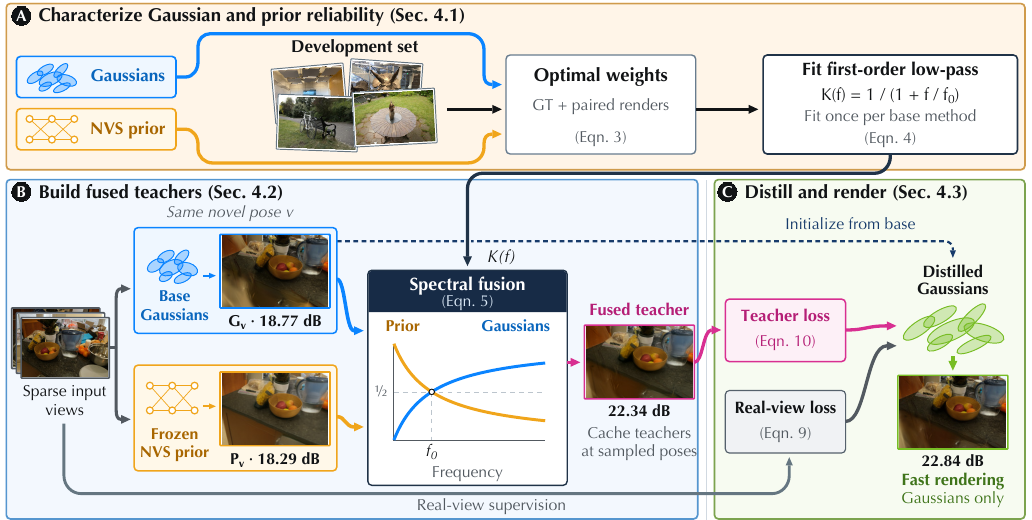}
  \caption{Overview of our method.
  (A) We fit the fusion kernel once on a development set.
  (B) The kernel fuses the prior render and the Gaussian render into a teacher.
  (C) The teachers and the real views supervise the Gaussians.}
  \label{fig:overview}
  \vspace{-10pt}
\end{figure*}

\Fig{overview} shows our method.
We first find, at every frequency, the optimal combination of the prior and the Gaussians, which gives the best result.
We model the combination as a weighted sum in the frequency domain, measure the optimal weights, and find that they follow a one-parameter law (\Sect{kernel}).
The weight defines a high-quality fused teacher (\Sect{teacher}).
To keep rendering fast, we finally distill it into the Gaussians (\Sect{distill}).

\begin{figure}[t]
  \centering
  \includegraphics[width=\linewidth]{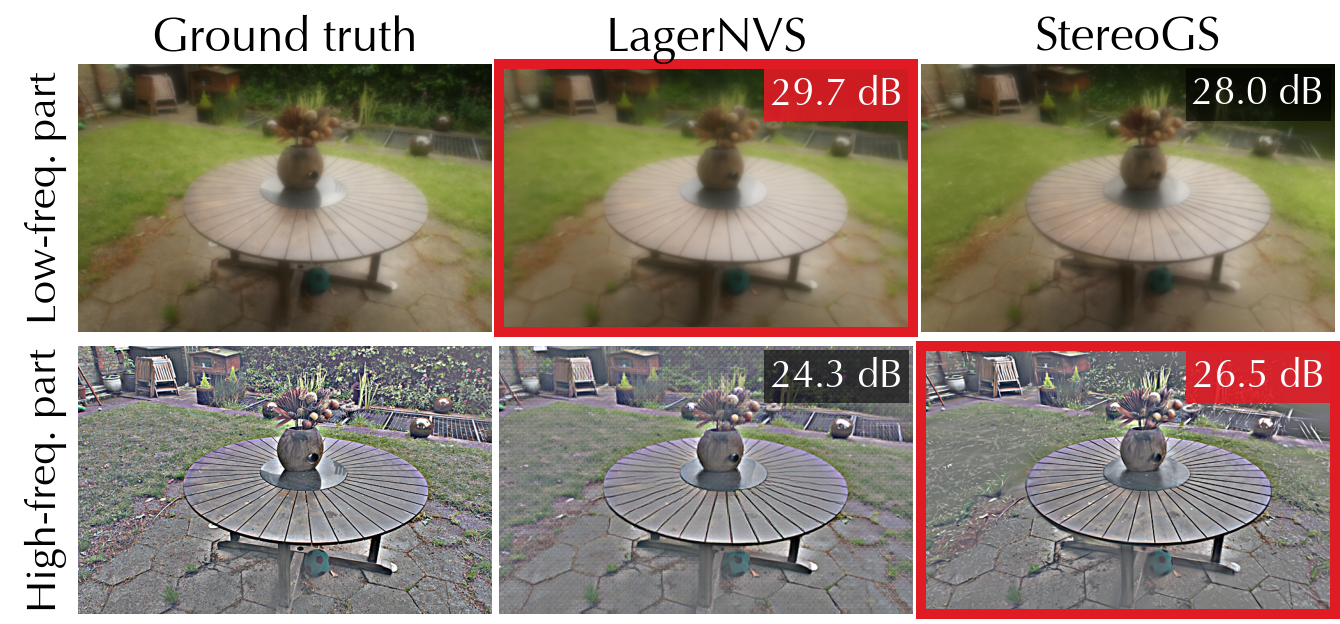}
  \caption{Comparing \lager and \stereogs on the low- and high-frequency parts using our first-order low-pass filter, on the \texttt{garden} scene with 12 input views.
  \lager indeed performs better at low frequencies, and \stereogs at high frequencies.}
  \label{fig:bands}
  \vspace{-10pt}
\end{figure}

\subsection{The Optimal Fusion Kernel}
\label{sec:kernel}

We show that the prior and the Gaussians are reliable in different frequency bands (\Sect{mot:pattern}).
Our goal is to find the best weight to combine them at every frequency.

Let $G_v$ and $P_v$ be the renders of the Gaussians and of the prior at a pose $v$, and $Y_v$ the ground-truth test image.
We combine the two renders as a weighted sum in the frequency domain, with a weight $w(f)$ on the prior and $1-w(f)$ on the Gaussians at each frequency $f$:
\begin{equation}
  \widehat{C_v}(\mathbf{f})=w(f)\,\widehat{P_v}(\mathbf{f})+\big(1-w(f)\big)\,\widehat{G_v}(\mathbf{f}),
  \label{eq:combine}
\end{equation}
where $\widehat{\cdot}$ is the Fourier transform, $\mathbf{f}$ is the 2D frequency and $f=\lvert\mathbf{f}\rvert$.
The optimal weight is the one that brings the combination closest to the ground truth:
\begin{equation}
  w^*(f)=\argmin_{w}\sum_{v}\big\lVert\widehat{Y_v}-\widehat{C_v}\big\rVert_f^2,
  \label{eq:argmin}
\end{equation}
where $\lVert\cdot\rVert_f$ sums over the Fourier coefficients at frequency $f$ and $v$ runs over the test views.
\Fig{optw} shows the result on five settings.
The optimal weight for prior is close to one at low frequencies and falls smoothly as the frequency increases, in every setting.
\Eqn{argmin} also has a closed-form solution.
We group the Fourier coefficients into 64 rings of equal width, and for the ring at $f$ we obtain
\begin{equation}
  w^*(f)=\frac{\sum_{v}\mathrm{Re}\left\langle \widehat{Y_v-G_v},\,\widehat{P_v-G_v}\right\rangle_f}{\sum_{v}\left\lVert\widehat{P_v-G_v}\right\rVert_f^2},
  \label{eq:wstar}
\end{equation}
where $\langle\cdot,\cdot\rangle_f$ is the inner product over the ring.

\paragraph{Parametrizing the optimal weight.}
The measured weight needs ground truth and is noisy, so it cannot be used directly at new scenes.
However, the optimal weight follows a similar pattern across scenes/settings (\Fig{optw}).
We therefore parameterize the optimal weight and fit it.
We compare different filter families (\Sect{ablation}), and find that a first-order low-pass filter with a single parameter fits the optimal weight well:
\begin{equation}
  K(f)=\frac{1}{1+f/f_0},
  \label{eq:kernel}
\end{equation}
where $f_0$ is the crossover frequency (weight=0.5).
It fits the measured optimum with a coefficient of determination $R^2$ of $0.993$ for \stereogs (\Fig{snr}), and of $0.989$ and $0.991$ for \dropg and \fsgs, using only one parameter.

\Fig{bands} applies the fitted kernel to the renders of \lager and \stereogs, which splits each of them into a low-frequency part and a high-frequency part.
\lager is indeed more accurate in the low-frequency part, and \stereogs in the high-frequency part.
Each wins its part by $2$\,dB{, which confirms our analysis and fitting}.

\subsection{The Fused Teacher}
\label{sec:teacher}

The kernel tells us how to combine the two models.
Given a prior model and a Gaussian model, we fit the kernel on the development set and use it to produce high-quality novel views on the test set.
We call the fused image a teacher $T_v$, because it later supervises the Gaussians.
At every 2D frequency $\mathbf{f}$, the teacher is the weighted sum of the two renders, with the kernel of \Eqn{kernel} as the weight on the prior:
\begin{equation}
  \widehat{T_v}(\mathbf{f})=K(\lvert\mathbf{f}\rvert)\,\widehat{P_v}(\mathbf{f})+\big(1-K(\lvert\mathbf{f}\rvert)\big)\,\widehat{G_v}(\mathbf{f}).
  \label{eq:teacher}
\end{equation}
$K$ is close to one at low frequencies and close to zero at high frequencies, so the teacher takes the low frequencies from the prior and the high frequencies from the Gaussians.
\Fig{bands} shows the two parts of one test view: \lager is more accurate in the low-frequency part, and \stereogs in the high-frequency part.
The only exception is the image mean, the coefficient at $\mathbf{f}=\mathbf{0}$.
\Eqn{kernel} gives it fully to the prior, but the prior drifts in global color on out-of-domain scenes.
{We set $K(0)=0$ to take it from the Gaussians.}

\subsection{Distillation into Gaussians}
\label{sec:distill}

The teacher needs the feed-forward model at render time, so it is slow.
To keep rendering fast, we distill the teacher into the Gaussians: we continue optimizing the Gaussians under the supervision of the teacher, and they then render alone.

\paragraph{Teacher poses.}
The Gaussians fail at views that the inputs do not cover, so we use the teacher to supervise them there.
The prior is reliable between the input cameras, so we sample the teacher poses there.
Let $\mathbf{c}_i$ and $\mathbf{q}_i$ be the center and the rotation of input camera $i$.
For each teacher pose $v$, we pick a random input camera $i$ and one of its nearest cameras $j$, and interpolate between them:
\begin{equation}
\begin{aligned}
  \mathbf{c}_v&=(1-t)\,\mathbf{c}_i+t\,\mathbf{c}_j+\boldsymbol{\epsilon},\\
  \mathbf{q}_v&=\mathrm{slerp}(\mathbf{q}_i,\mathbf{q}_j,t),
\end{aligned}
  \label{eq:pose}
\end{equation}
where $t$ is uniform in $[0.15,0.85]$ and $\mathrm{slerp}$ is the spherical interpolation of rotations.
The noise $\boldsymbol{\epsilon}\sim\mathcal{N}(\mathbf{0},(0.1\,\bar d)^2\mathbf{I})$ moves the pose off the line between the two cameras and improves the coverage of the teacher, with $\bar d$ set to the mean distance between input cameras.
We sample $M=512$ poses and render the teacher once per pose.

\paragraph{Balancing real and teacher views.}
The teacher is not perfect, and the input images are the only error-free signal, with true high-frequency detail and true color.
We therefore keep the real images in the supervision.
Each optimization step draws one input image $I_n$ and one teacher image $T_v$ uniformly at random:
\begin{equation}
  n\sim\mathcal{U}\{1,\dots,N\},\qquad v\sim\mathcal{U}\{1,\dots,M\},
  \label{eq:sample}
\end{equation}
where $N$ is the number of input views.
The two images enter the loss with weights $\lambda_r$ and $\lambda_t$:
\begin{equation}
  \mathcal{L}=\lambda_r\,\mathcal{L}_{\text{real}}+\lambda_t\,\mathcal{L}_{\text{teacher}}.
  \label{eq:loss}
\end{equation}
{We balance the two sources with the ratio $\lambda_r:\lambda_t=1:1$.}

\paragraph{Loss.}
We now define the two terms of \Eqn{loss}.
Both terms use the standard 3DGS photometric loss~\cite{kerbl20233d}, $\ell=0.8\,\ell_1+0.2\,(1-\mathrm{SSIM})$.
Let $\hat I_n$ and $\hat I_v$ be the renders of the Gaussians at the poses of $I_n$ and $T_v$.
The input image is real, so the first term compares the render with it directly:
\begin{equation}
  \mathcal{L}_{\text{real}}=\ell\big(\hat I_n,\,I_n\big).
  \label{eq:lreal}
\end{equation}
The global color of the teacher image is less accurate, so the teacher does not supervise the image mean.
The second term subtracts the per-channel mean $\Mean(\cdot)$ from both the render and the teacher before the loss:
\begin{equation}
  \mathcal{L}_{\text{teacher}}=\ell\big(\hat I_v-\Mean(\hat I_v),\,T_v-\Mean(T_v)\big).
  \label{eq:lteacher}
\end{equation}
The global color is then learned from the input images only.

\section{Experimental Setup}
\label{sec:setup}

\paragraph{Datasets.}
We use LLFF~\cite{mildenhall2019llff} with 3, 6 and 9 input views over 8 scenes, and Mip-NeRF 360~\cite{barron2022mipnerf360} with 12 and 24 input views over 9 scenes, for 42 settings in total.
A setting is one scene at one input view count, written as \texttt{Garden-12}.

We use the image resolution of the official setting of each base method.
The training and test poses follow the standard sparse-view split of these datasets~\cite{niemeyer2022regnerf,zhu2024fsgs}.

\paragraph{Implementation details.}
We use the official general \lager checkpoint~\cite{szymanowicz2026lagernvs} as the NVS prior.
Following its official setting, we resize the input images to a long side of 512 pixels before feeding them to the prior.
The prior also renders at a long side of 512 pixels, and we resample its renders to the evaluation resolution.

We fit the crossover frequency $f_0$ of the first-order low-pass filter (\Sect{kernel}) on development settings, once per base method.
To evaluate every setting with a crossover fitted without it, we use two disjoint development sets.
Each set has five settings, one per view count.

\begin{table*}[t]
  \centering
  \footnotesize
  \setlength{\tabcolsep}{2.2pt}
  \resizebox{\linewidth}{!}{\begin{tabular}{lccc|ccc|ccc|ccc|ccc|ccc}
    \toprule
    & \multicolumn{9}{c|}{LLFF (8 scenes)} & \multicolumn{6}{c|}{Mip-NeRF 360 (9 scenes)} & \multicolumn{3}{c}{Overall (42 settings)} \\
    & \multicolumn{3}{c|}{3-view} & \multicolumn{3}{c|}{6-view} & \multicolumn{3}{c|}{9-view} & \multicolumn{3}{c|}{12-view} & \multicolumn{3}{c|}{24-view} & \multicolumn{3}{c}{} \\
    Method & PSNR & SSIM & LPIPS & PSNR & SSIM & LPIPS & PSNR & SSIM & LPIPS & PSNR & SSIM & LPIPS & PSNR & SSIM & LPIPS & PSNR & SSIM & LPIPS \\
    \midrule
    \rowcolor{rowlit} \multicolumn{19}{l}{\textit{Other Gaussians}} \\
    3DGS~\cite{kerbl20233d} & 16.02 & .465 & .378 & 19.45 & .627 & .268 & 21.13 & .715 & .214 & 17.34 & .454 & .405 & 21.74 & .662 & .248 & 19.16 & .583 & .304 \\
    DNGaussian~\cite{li2024dngaussian} & 19.12 & .591 & .294 & 22.18 & .755 & .198 & 23.17 & .788 & .180 & -- & -- & -- & -- & -- & -- & -- & -- & -- \\
    CoR-GS~\cite{zhang2024corgs} & 20.45 & .712 & .196 & 24.49 & .837 & .115 & 26.06 & .874 & .089 & 17.40 & .488 & .439 & 22.23 & .673 & .282 & 22.02 & .710 & .231 \\
    Binocular3DGS~\cite{han2024binocular} & 21.44 & .751 & .168 & 24.87 & .845 & .106 & 26.17 & .877 & .090 & -- & -- & -- & -- & -- & -- & -- & -- & -- \\
    \midrule
    \rowcolor{rowlit} \multicolumn{19}{l}{\textit{Feed-forward Methods}} \\
    \anysplat~\cite{jiang2025anysplat} & 11.03 & .322 & .419 & 12.84 & .457 & .309 & 13.21 & .528 & .279 & 10.44 & .261 & .556 & 14.50 & .316 & .467 & 12.41 & .373 & .411 \\
    \depthsplat~\cite{xu2025depthsplat} & 14.38 & .383 & .456 & 14.92 & .395 & .462 & 14.61 & .393 & .477 & 17.12 & .410 & .446 & 18.29 & .457 & .393 & 15.95 & .409 & .445 \\
    LVSM~\cite{jin2025lvsm} & 20.15 & .642 & .201 & 22.44 & .727 & .144 & 23.13 & .760 & .126 & 15.90 & .320 & .571 & 16.99 & .332 & .545 & 19.57 & .545 & .329 \\
    \lager~\cite{szymanowicz2026lagernvs} & 21.41 & .637 & .193 & 22.94 & .692 & .155 & 23.19 & .706 & .146 & 20.63 & .487 & .339 & 21.96 & .528 & .287 & 21.99 & .605 & .228 \\
    \midrule
    \rowcolor{rowlit} \multicolumn{19}{l}{\textit{Diffusion-guided Gaussians}} \\
    Difix3D+~\cite{wu2025difix3d} & 18.07 & .564 & .259 & 21.27 & .708 & .181 & 22.49 & .760 & .154 & 15.67 & .339 & .441 & 19.08 & .515 & .322 & 19.22 & .570 & .277 \\
    \stereogs + Difix3D+~\cite{wu2025difix3d} & 20.28 & .685 & .183 & 22.85 & .769 & .138 & 23.65 & .798 & .124 & 17.99 & .470 & .351 & 21.08 & .619 & .248 & 21.09 & .662 & .213 \\
    \midrule
    \rowcolor{rowlit} \multicolumn{19}{l}{\textit{Sparse-view Gaussians with Ours}} \\
    \stereogs~\cite{yuan2026stereogs} & 21.99 & .773 & .156 & 25.26 & \underline{.854} & \underline{.109} & 26.14 & \underline{.876} & \underline{.096} & 19.60 & .565 & .370 & 23.21 & \underline{.729} & .236 & 23.15 & \underline{.754} & .199 \\
    \rowcolor{rowteacher} StereoGS + Ours-Teacher & \textbf{23.19} & \underline{.777} & \underline{.155} & \textbf{26.04} & .852 & .110 & \underline{26.71} & .872 & .099 & \textbf{21.64} & \underline{.571} & \underline{.357} & \textbf{24.57} & .718 & \underline{.235} & \textbf{24.37} & .753 & \underline{.196} \\
    \rowcolor{rowdistill} StereoGS + Ours-Distilled & \underline{22.60} & \textbf{.778} & \textbf{.149} & \underline{25.99} & \textbf{.860} & \textbf{.099} & \textbf{26.90} & \textbf{.883} & \textbf{.085} & \underline{21.15} & \textbf{.589} & \textbf{.346} & \underline{24.47} & \textbf{.743} & \textbf{.215} & \underline{24.16} & \textbf{.766} & \textbf{.184} \\
    \midrule
    \dropg~\cite{park2025dropgaussian} & 20.49 & .708 & .200 & 24.56 & \underline{.836} & \underline{.117} & 25.99 & \textbf{.874} & \underline{.090} & 18.45 & .507 & .420 & 22.65 & \underline{.686} & .286 & 22.34 & .716 & .229 \\
    \rowcolor{rowteacher} DropGaussian + Ours-Teacher & \textbf{22.34} & \textbf{.727} & \textbf{.190} & \textbf{25.59} & \underline{.836} & \underline{.117} & \underline{26.49} & .866 & .094 & \textbf{21.30} & \textbf{.533} & \textbf{.392} & \textbf{24.26} & .682 & \textbf{.276} & \textbf{23.94} & \underline{.723} & \textbf{.220} \\
    \rowcolor{rowdistill} DropGaussian + Ours-Distilled & \underline{21.44} & \underline{.721} & \textbf{.190} & \underline{25.29} & \textbf{.840} & \textbf{.111} & \textbf{26.50} & \textbf{.874} & \textbf{.086} & \underline{20.07} & \underline{.530} & \underline{.405} & \underline{23.81} & \textbf{.696} & \textbf{.276} & \underline{23.35} & \textbf{.727} & \textbf{.220} \\
    \midrule
    \fsgs~\cite{zhu2024fsgs} & 20.51 & .700 & .203 & 24.21 & .824 & .127 & 25.58 & \underline{.863} & \underline{.100} & 17.90 & .476 & .430 & 22.28 & \underline{.664} & .294 & 22.00 & .699 & .237 \\
    \rowcolor{rowteacher} FSGS + Ours-Teacher & \textbf{22.30} & \textbf{.722} & \textbf{.190} & \textbf{25.26} & \underline{.825} & \underline{.125} & \underline{26.07} & .855 & .102 & \textbf{20.88} & \textbf{.511} & \textbf{.396} & \textbf{23.89} & \underline{.664} & \textbf{.282} & \textbf{23.62} & \underline{.709} & \textbf{.225} \\
    \rowcolor{rowdistill} FSGS + Ours-Distilled & \underline{21.52} & \underline{.718} & \underline{.191} & \underline{25.15} & \textbf{.834} & \textbf{.117} & \textbf{26.24} & \textbf{.868} & \textbf{.091} & \underline{19.87} & \underline{.509} & \underline{.410} & \underline{23.49} & \textbf{.677} & \underline{.284} & \underline{23.18} & \textbf{.715} & \textbf{.225} \\
    \bottomrule
  \end{tabular}}
  \caption{Results by dataset and view count.
  Bold and underline mark the best and second best per base method.}
  \label{tab:main}
  \vspace{-10pt}
\end{table*}

The first set includes \texttt{Fern-3}, \texttt{Room-6} and \texttt{Horns-9} from LLFF, and \texttt{Garden-12} and \texttt{Bicycle-24} from Mip-NeRF 360, and its $f_0$ is used for the other 37 settings.
The second set is sampled in the same way from the other scenes and includes \texttt{Leaves-3}, \texttt{Fortress-6}, \texttt{Trex-9}, \texttt{Stump-12} and \texttt{Bonsai-24}, and its fitted $f_0$ is used for the five settings of the first set.
All 42 settings are therefore held out from the fit that evaluates them.
For distillation, we sample $M=512$ teacher poses and render each teacher once.
We then optimize the Gaussians of the base model for 12{,}000 steps with Adam, with $\lambda_r=\lambda_t=1$ and no densification.
Distilling one scene takes about three minutes on RTX5090.
\Supp{impl} gives the remaining details.

\paragraph{Baselines.}
Our method is plug-and-play, so we add it to per-scene sparse-view methods and compare each enhanced model against its base method.
These methods optimize the Gaussians of each scene from its sparse inputs.
We use \stereogs~\cite{yuan2026stereogs}, \dropg~\cite{park2025dropgaussian} and \fsgs~\cite{zhu2024fsgs}.

We also compare with two other kinds of methods:
\begin{itemize}
  \item \textbf{Feed-forward methods} predict the novel views or the Gaussians from the sparse inputs in one pass. We use \lager~\cite{szymanowicz2026lagernvs}, \lvsm~\cite{jin2025lvsm}, \depthsplat~\cite{xu2025depthsplat} and \anysplat~\cite{jiang2025anysplat}.
  \item \textbf{Diffusion refinement} enhances the Gaussian renders with a diffusion model. We use \difix~\cite{wu2025difix3d}, with the Gaussian model of its own and on top of \stereogs.
\end{itemize}

We report PSNR, SSIM~\cite{wang2004ssim} and LPIPS~\cite{zhang2018lpips}.

\paragraph{Variants.} \teacher is the fused teacher (\Sect{teacher}), and \distilled is the distilled Gaussians (\Sect{distill}).

\section{Evaluation}
\label{sec:eval}

We first show that our method improves the quality of every base method and achieve state-of-the-art result (\Sect{results}), which is also visible in the renders (\Sect{qual}).
We then show that \distilled renders as fast as its base model (\Sect{speed}).
We ablate the kernel and the components of the distillation (\Sect{ablation}) and analyze further (\Sect{analysis}).
{We finally compare \lager with four other priors} (\Sect{prior}).

\subsection{Quantitative Results}
\label{sec:results}

Our method improves every base method.
\Tbl{main} compares \stereogs, \dropg and \fsgs with our two variants on every dataset and view count.
Overall, \teacher improves them by $1.22$, $1.60$ and $1.62$\,dB in PSNR, and \distilled by $1.01$, $1.01$ and $1.18$\,dB, respectively.
This shows that our plug-and-play method effectively improves sparse-view Gaussian models.
\stereogs with our method also achieves state-of-the-art results against all other methods.
In contrast, diffusion refinement with \difix lowers the PSNR of \stereogs, which shows that using a prior without selecting what to take from it can lower the quality of sparse-view Gaussians.

\distilled even exceeds \teacher in some metrics: it has the better SSIM overall for all three base methods, and a better or equal LPIPS.
This is because the distillation uses the real images in addition to the teacher images, so the student can find a solution between the two that is better than the teacher.
Note that \distilled also renders $28\times$ faster than \teacher overall (\Sect{speed}).

\begin{figure*}[t]
  \centering
  \includegraphics[width=\linewidth]{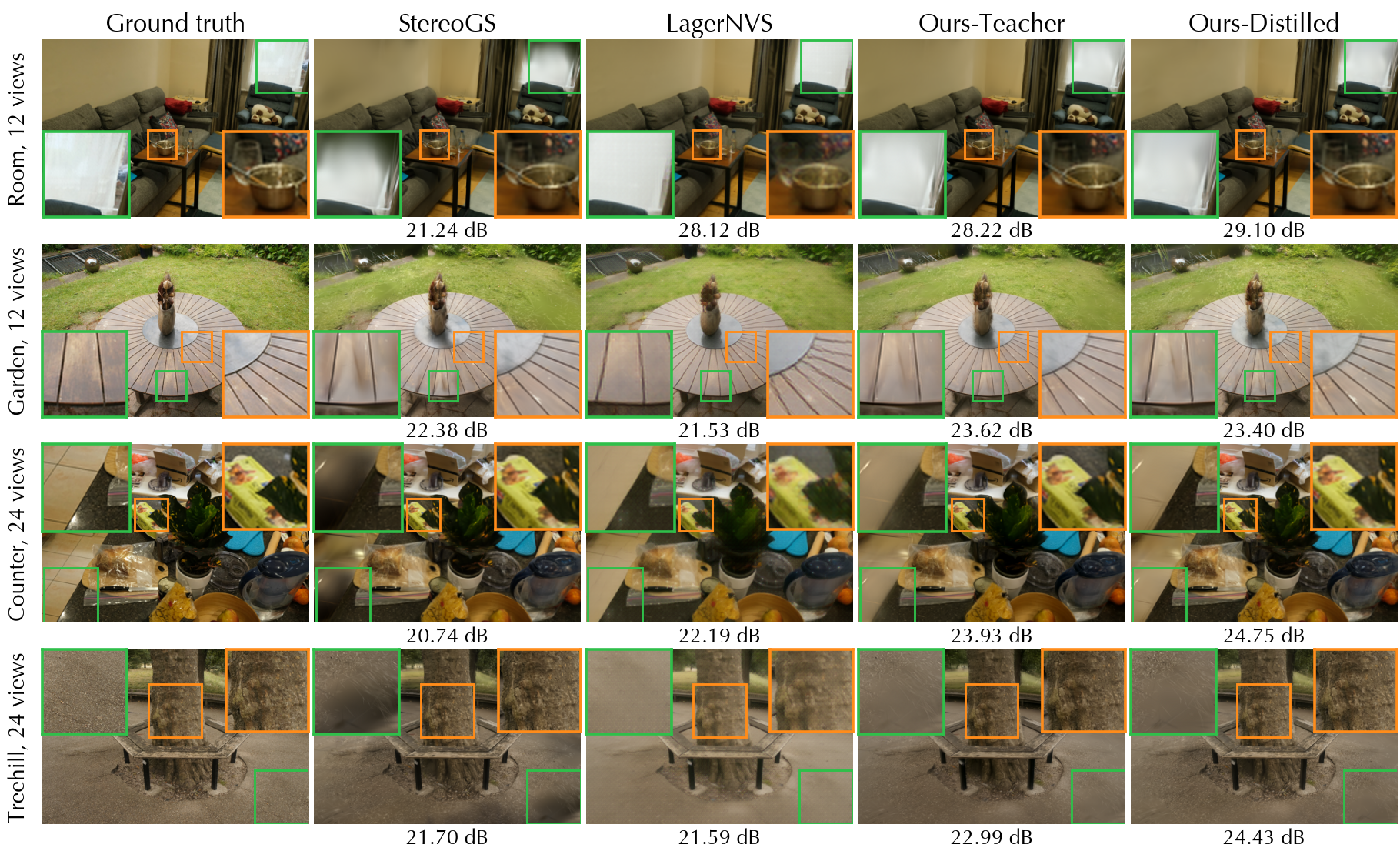}
  \caption{Qualitative comparison on \stereogs, with the PSNR under each image.
  {Green boxes mark where \stereogs fails, orange boxes where \lager fails, and each box is enlarged in a corner of its image.}}
  \label{fig:qualitative}
\end{figure*}
\begin{figure*}[t]
  \centering
  \begin{minipage}[t]{0.48\textwidth}
    \centering
    \includegraphics[width=\linewidth]{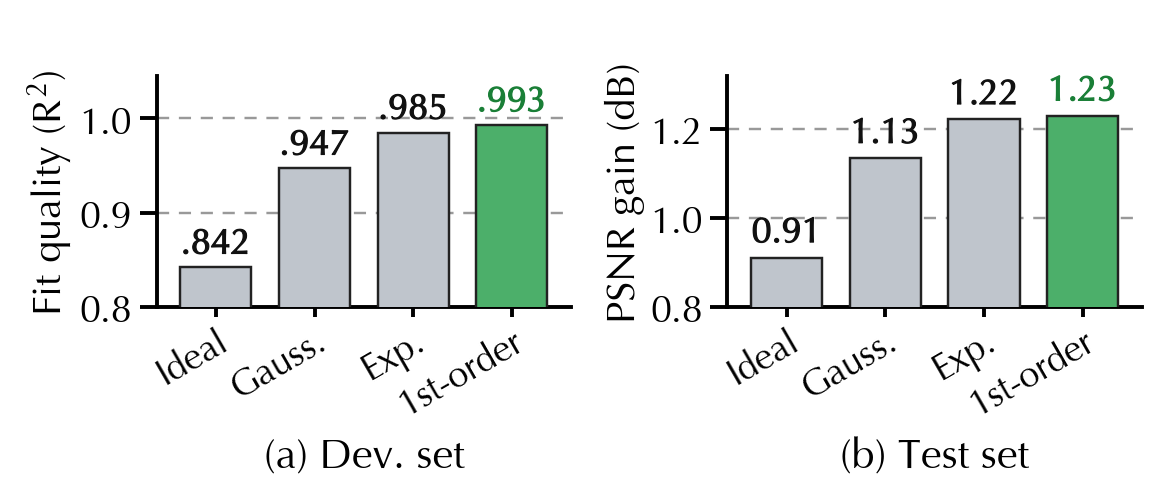}
    \caption{Kernels with one parameter.
    (a) Fit quality on the development set.
    (b) PSNR gain of \teacher on the test set.}
    \label{fig:snr}
  \end{minipage}\hfill
  \begin{minipage}[t]{0.48\textwidth}
    \centering
    \includegraphics[width=\linewidth]{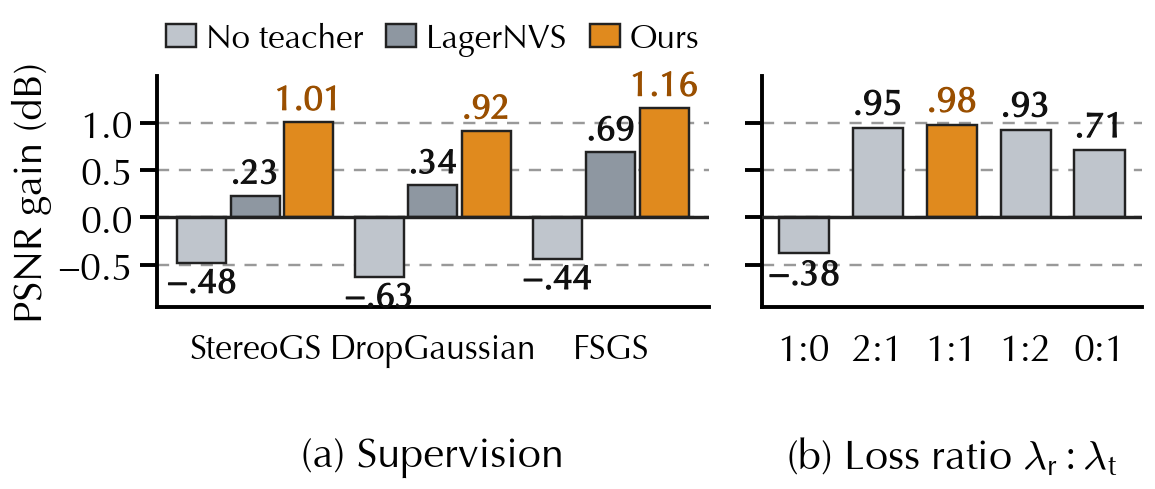}
    \caption{Ablation of the distillation, as the PSNR gain of \distilled over its base method.}
    \label{fig:ablation}
  \end{minipage}
  \vspace{-5pt}
\end{figure*}

\subsection{Qualitative Results}
\label{sec:qual}

\Fig{qualitative} visualizes the renders of \stereogs, \lager and our two variants.
\stereogs is sharp but carries smears, floaters and distorted structures at views far from the inputs (green insets).
\lager gives the correct layout, but its renders are blurred and carry grid artifacts (orange insets).
Our variants take the layout from the prior and the detail from the Gaussians, so they are sharper than \lager and cleaner than \stereogs.
{See \Supp{qual} for more examples.}

\subsection{Rendering Speed}
\label{sec:speed}

With our method, Gaussian splatting keeps its fast rendering speed.
\Tbl{fps} gives the rendering speed and the quality of \stereogs, \lager and our two variants, averaged over all 42 settings.
\distilled renders at $917$\,FPS on an RTX 5090, against $882$\,FPS for \stereogs: distillation adds no Gaussians, so it does not slow down rendering.
\teacher runs \lager for every frame and renders at $33$\,FPS.
This supports our choice to distill: \distilled is $28\times$ faster than \teacher at a similar quality.
\Fig{teaser} shows the trade-off between speed and quality of all methods, and our method gives the best trade-off for every dataset and view count.
The cost of our method is paid once: distilling one scene takes about three minutes, which adds $6\%$ to $8\%$ to the training time of \stereogs.

\begin{table}[t]
  \centering
  \footnotesize
  \setlength{\tabcolsep}{3.5pt}
  \begin{tabular}{lcccc}
    \toprule
    Method & FPS$\uparrow$ & PSNR$\uparrow$ & SSIM$\uparrow$ & LPIPS$\downarrow$ \\
    \midrule
    \lager & 34.7 & 21.99 & .605 & .228 \\
    \stereogs & \underline{882} & 23.15 & \underline{.754} & .199 \\
    \rowcolor{rowteacher} StereoGS + Ours-Teacher & 33.1 & \textbf{24.37} & .753 & \underline{.196} \\
    \rowcolor{rowdistill} StereoGS + Ours-Distilled & \textbf{917} & \underline{24.16} & \textbf{.766} & \textbf{.184} \\
    \bottomrule
  \end{tabular}
  \caption{Rendering speed and quality, averaged over all 42 settings.}
  \label{tab:fps}
  \vspace{-10pt}
\end{table}
\subsection{Ablation Studies}
\label{sec:ablation}

\paragraph{Choice of the kernel.}
We choose the kernel by two criteria: it must fit the measured optimal weight well, and it must have as few parameters as possible, so that a fit on the development set generalizes.
\Fig{snr}(a) plots the fit quality of the four families with one parameter on the first development set, and \Fig{snr}(b) the gain of \teacher on the 37 test settings.
The first-order kernel fits the best and gains the most.
A Gaussian blur and a hard cutoff discard the prior too early, and fit clearly worse.
Families with two or three parameters reach slightly better results, but about the same as the first-order kernel.
We therefore choose the first-order kernel for its simplicity.
See \Supp{analysis} for more kernel families and corresponding evaluation.

\paragraph{Teacher.}
The fused teacher is necessary.
\Fig{ablation}(a) compares three ways to supervise.
Without a teacher, the extra steps on the real images lower the PSNR of every base method due to overfitting, so the gain does not come from longer optimization.
Using only \lager as the teacher, the gain is smaller for every base method, because the Gaussians also learn the high-frequency artifacts of the prior.

\paragraph{Balancing real and teacher views.}
Both losses are necessary, and their ratio needs no tuning.
\Fig{ablation}(b) varies the ratio $\lambda_r:\lambda_t$.
Either loss alone is worse.
The gain is almost the same between $2:1$ and $1:2$, so we use $1:1$.

\subsection{Further Analysis}
\label{sec:analysis}

\paragraph{Generality of the frequency pattern.}
The pattern of \Sect{motivation} holds over all 42 settings: on average, the prior is more accurate than \stereogs at low frequencies and less accurate at high frequencies, and the two curves cross once.
See \Supp{analysis} for the pattern.
A weaker Gaussian method has a larger fitted $f_0$: $0.041$ for \stereogs, $0.062$ for \dropg and $0.077$ for \fsgs.

\begin{table}[t]
  \centering
  \footnotesize
  \setlength{\tabcolsep}{3.5pt}
  \begin{tabular}{lcccc}
    \toprule
    Method & Steps & PSNR$\uparrow$ & SSIM$\uparrow$ & LPIPS$\downarrow$ \\
    \midrule
    StereoGS & 30k & 23.15 & \underline{.754} & .199 \\
    StereoGS, scaled schedule & 42k & 23.29 & \underline{.754} & .201 \\
    \rowcolor{rowteacher} StereoGS + Ours-Teacher & 30k & \textbf{24.37} & .753 & \underline{.196} \\
    \rowcolor{rowdistill} StereoGS + Ours-Distilled & 30k + 12k & \underline{24.16} & \textbf{.766} & \textbf{.184} \\
    \bottomrule
  \end{tabular}
  \caption{\stereogs trained as long as \distilled.}
  \label{tab:budget}
  \vspace{-5pt}
\end{table}

\begin{figure}[t]
  \centering
  \includegraphics[width=\linewidth]{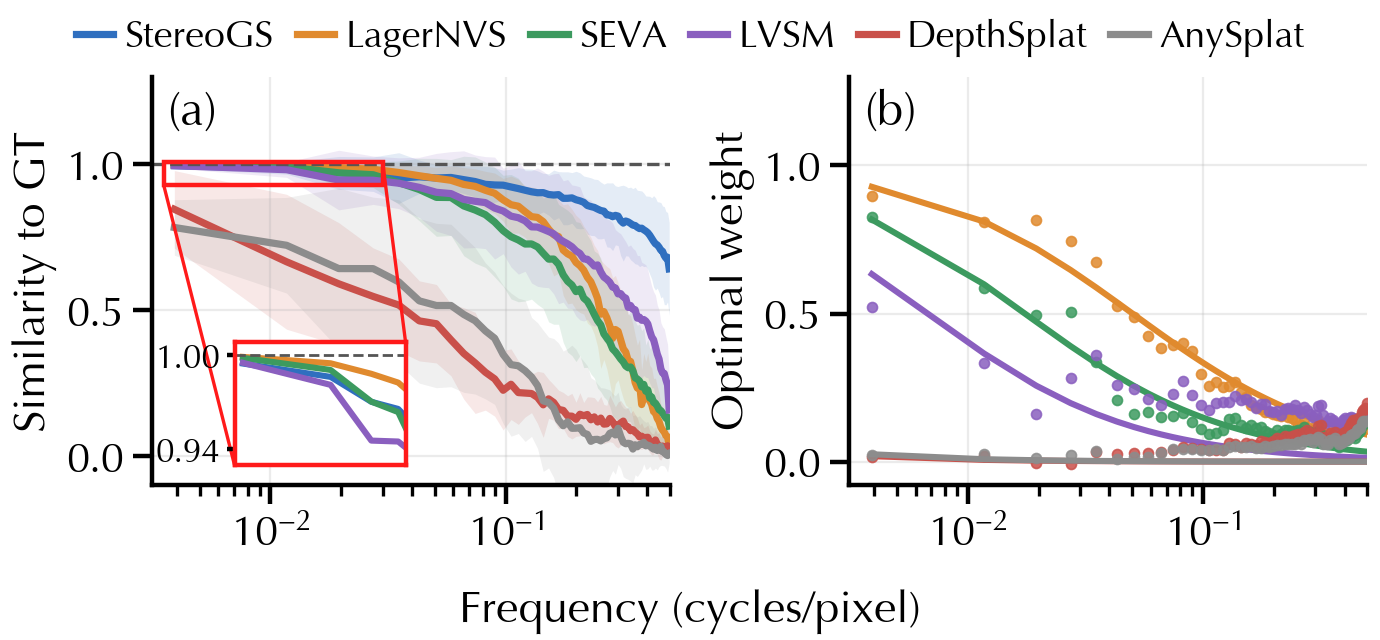}
  \caption{Five priors on LLFF.
  (a) Similarity to GT per frequency.
  (b) Optimal weight against \stereogs (markers) and its fit (lines).}
  \label{fig:priorcurve}
  \vspace{-5pt}
\end{figure}

\paragraph{Sensitivity to $f_0$.}
Our method is insensitive to the crossover.
The gain of \teacher stays within $0.05$\,dB of its best while $f_0$ varies from $0.62$ to $1.41$ times its fitted value, for all three base methods (\Supp{analysis}).
It is also insensitive to the development set: when we refit $f_0$ on 200 random sets of five settings, the gain of \teacher on the remaining settings stays above $1.0$\,dB for all three base methods in $95\%$ of the draws.
The crossover even transfers across base methods: with the $f_0$ of \stereogs, \dropg and \fsgs keep more than $95\%$ of their gain.
It also transfers across datasets: an $f_0$ fitted on LLFF alone loses at most $0.03$\,dB on Mip-NeRF 360.

\paragraph{Equal training budget.}
The gain does not come from the extra optimization steps.
Our distillation adds 12{,}000 steps to the 30{,}000 steps of \stereogs.
We therefore train \stereogs for 42{,}000 steps, with its learning-rate schedule and every step-dependent setting scaled accordingly, and compare it with our two variants (\Tbl{budget}).
The longer training gains only $0.14$\,dB, while \distilled gains $1.01$\,dB.

\subsection{Choice of the Prior}
\label{sec:prior}
{We also test \seva~\cite{zhou2025seva}, \lvsm~\cite{jin2025lvsm}, \depthsplat and \anysplat as the prior on LLFF.
\lager, \seva and \lvsm render low frequencies accurately (\Fig{priorcurve}(a)), so their optimal weight there exceeds $0.5$ (\Fig{priorcurve}(b)) and \teacher improves \stereogs by $0.29$ to $0.83$\,dB (\Tbl{priors}).
\depthsplat and \anysplat are less accurate at every frequency, so their $f_0$ is almost zero.}
All priors have a lower PSNR than \stereogs, yet three of them improve it.
This shows that with a suitable fusion, even a weaker prior can help per-scene Gaussians.
{\lager has both the highest PSNR on its own and the largest gain, so we use it as our prior.}
See \Supp{prior} for more details.

\begin{table}[t]
  \centering
  \footnotesize
  \setlength{\tabcolsep}{3.5pt}
  \begin{tabular}{lcccc}
    \toprule
    Prior & PSNR alone & $f_0$ & $R^2$ & \teacher $\Delta$PSNR \\
    \midrule
    \anysplat & 12.36 & 0.0001 & 0.39 & $-$0.04 \\
    \depthsplat & 14.64 & 0.0001 & 0.11 & $-$0.02 \\
    LVSM & \underline{21.91} & 0.0067 & 0.81 & +0.29 \\
    SEVA & 20.36 & 0.0175 & \underline{0.98} & \underline{+0.41} \\
    \lager & \textbf{22.51} & 0.0501 & \textbf{0.99} & \textbf{+0.83} \\
    \bottomrule
  \end{tabular}
  \vspace{-3pt}
  \caption{Different priors fused into \stereogs on LLFF, where \stereogs alone reaches 24.46\,dB. {$\Delta$PSNR is the gain of \teacher over \stereogs.}}
  \label{tab:priors}
\end{table}

\section{Conclusion}
\label{sec:conclusion}

{We enhance sparse-view Gaussians with a feed-forward NVS prior, which renders low image frequencies more accurately than the Gaussians and high ones less accurately.
To decide the optimal spectral fusion between them,
a first-order low-pass kernel is fitted once on a development set, which is used to fuse the two renders into a teacher that we distill into the Gaussians.
{It improves different Gaussians, renders at $917$\,FPS, and works with different priors.}}

\begin{figure}[t]
  \centering
  \includegraphics[width=\linewidth]{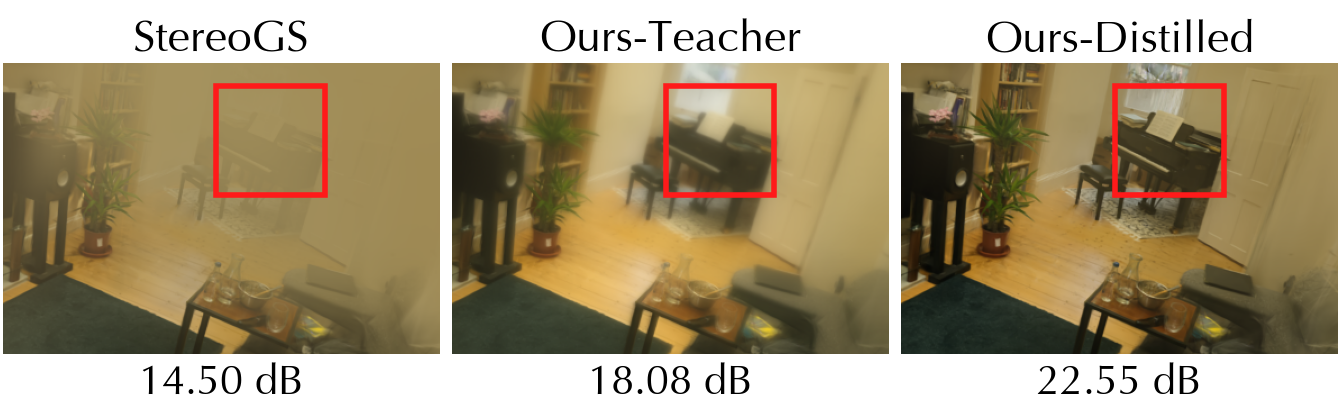}
  \caption{A failure case on \texttt{room} with 24 views: floaters of \stereogs partly remain in our results.}
  \label{fig:failure}
\end{figure}

\paragraph{Limitations.}
\label{sec:limits}
{Our method helps only if the prior renders low frequencies {more} accurately than the Gaussians: with \depthsplat or \anysplat, due to their low quality, the fitted $f_0$ is about $10^{-4}$ cycles per pixel and the teacher can't improve the PSNR (\Sect{prior}).}
Moreover, when the base model is trained very poorly, our method reduces its {artifacts} but does not fully remove them (\Fig{failure}).

\clearpage

{
    \small
    \bibliographystyle{ieeenat_fullname}
    \bibliography{main}
}

\clearpage
\appendix
\newpage
\setcounter{figure}{0}\setcounter{table}{0}
\renewcommand{\thefigure}{S\arabic{figure}}\renewcommand{\thetable}{S\arabic{table}}
\renewcommand{\theHfigure}{S\arabic{figure}}\renewcommand{\theHtable}{S\arabic{table}}
\twocolumn[{\centering
  {\Large\bf \thetitle\par}\vspace{0.5em}{\Large Supplementary Material\par}\vspace{1.0em}
  \includegraphics[width=\linewidth]{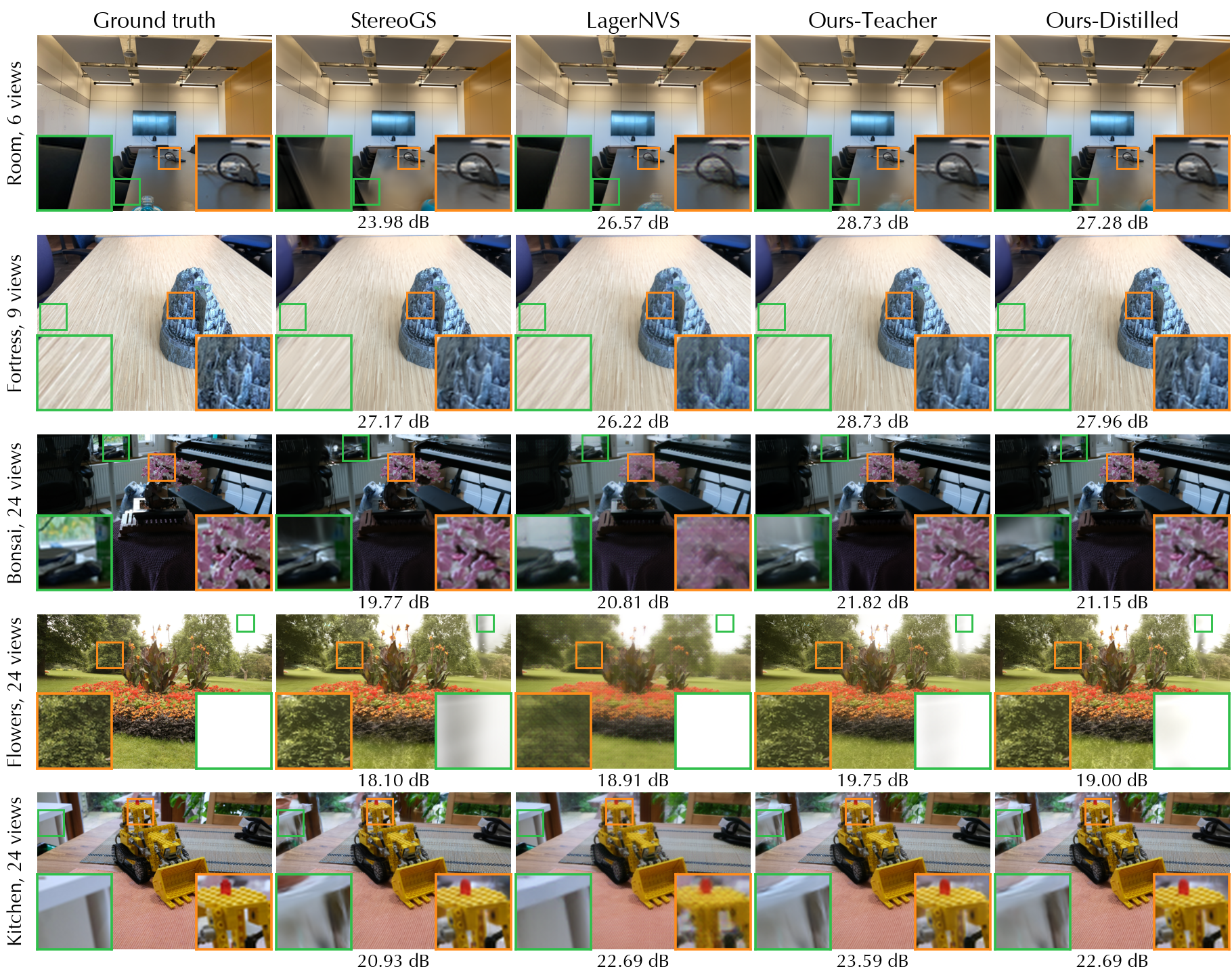}
  \captionof{figure}{More test views on \stereogs, as in \Fig{qualitative}.
  Green boxes mark where \stereogs fails, orange boxes where \lager fails.}
  \label{fig:qualsupp}\vspace{1.2em}
}]

\section{More Qualitative Results}
\label{sec:supp:qual}
\Fig{qualsupp} shows five more test views, two of them on Mip-NeRF 360 scenes outside both development sets.

\section{Results of Unseen Scenes}
\label{sec:supp:groups}
\label{sec:supp:disjoint}

On all 42 settings, \distilled raises the PSNR in 41 for \stereogs and in  42 for \dropg and \fsgs.
16 of the 42 settings are on scenes that appear in neither development set: Flower and Orchids of LLFF, and Counter, Flowers, Kitchen, Room and Treehill of Mip-NeRF 360.
On all sixteen, both variants raise the PSNR of every base method, by $1.17$ to $2.08$\,dB on average (\Tbl{disjoint}).
This shows that our method generalizes to unseen scenes.

\begin{table}[H]
  \centering
  \footnotesize
  \setlength{\tabcolsep}{4pt}
  \begin{tabular}{lcc}
    \toprule
    Base & \teacher & \distilled \\
    \midrule
    \stereogs & $+1.47$ & $+1.17$ \\
    \dropg & $+2.02$ & $+1.27$ \\
    \fsgs & $+2.08$ & $+1.44$ \\
    \bottomrule
  \end{tabular}
  \caption{$\Delta$PSNR over the base method on the sixteen settings whose scene is in neither development set.}
  \label{tab:disjoint}
\end{table}

\begin{figure*}[t]
  \centering
  \includegraphics[width=\textwidth]{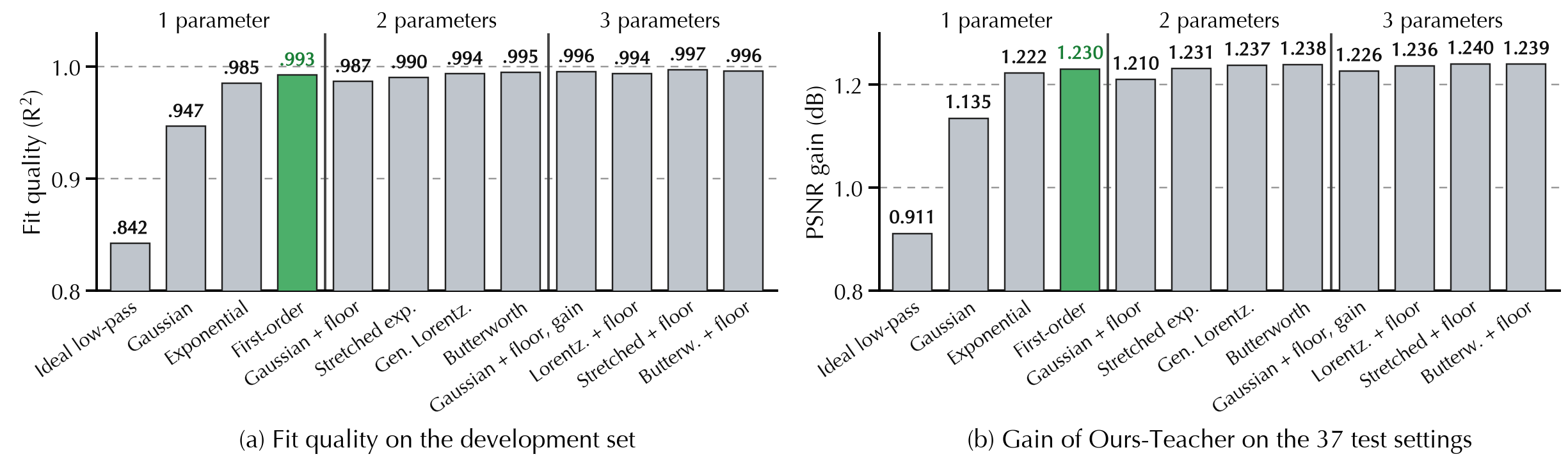}
  \caption{Kernels by number of parameters: (a) fit quality on the first development set, (b) gain of \teacher on the other 37.}
  \label{fig:kernel_all}
\end{figure*}

\section{Implementation Details}
\label{sec:supp:impl}

\paragraph{Fitting $f_0$.}
\label{sec:supp:dev}
We fit $f_0$ per base method on each of the two development sets of \Sect{setup} by least squares against the measured optimal weight $w^*$, weighted by the residual power per ring.
The fitted crossovers on the first set are $f_0=0.0405$, $0.0623$ and $0.0769$ cycles per pixel for \stereogs, \dropg and \fsgs, and on the second they are $0.0459$, $0.0770$ and $0.0862$.

\paragraph{Learning rates.}
The position learning rate decays exponentially from $1.6\times10^{-4}$ to $1.6\times10^{-6}$, scaled by the scene extent.
The other learning rates are $10^{-3}$ for scaling and spherical harmonic DC, $5\times10^{-4}$ for rotation, $10^{-2}$ for opacity, and $5\times10^{-5}$ for higher-order harmonics.

\paragraph{Base methods and prior.}
We run the official code and settings of \stereogs, \dropg and \fsgs.
\lager is conditioned on the input images of the base method at a long side of 512 pixels and renders on a canvas with the aspect ratio of its inputs.
We resample its renders to the evaluation grid.
The evaluation resolution is that of the official pipelines, which downsample the images by eight: $504\times378$ on LLFF, about $390\times260$ on the indoor scenes of Mip-NeRF 360 and about $630\times415$ on its outdoor scenes.

\paragraph{Training overhead.}
\teacher adds one Gaussian render and one FFT fusion of 1 to 2\,ms to the \lager render.
On one RTX 5090, distillation takes $2.5$ minutes on Fern-3 and $2.9$ minutes on Garden-12, against $31.6$ and $45.3$ minutes for the training of \stereogs.

\paragraph{Feed-forward and diffusion baselines.}
Each feed-forward baseline runs one forward pass on the input images of \stereogs, and we score its renders at our evaluation resolution.
\depthsplat uses the official checkpoint trained on RealEstate10K and DL3DV at its training width of 768 pixels, and at 512 pixels for the settings with 24 views, which do not fit into 32\,GB of memory.
\anysplat follows its release evaluation protocol: it builds the Gaussians from the input views and predicts the cameras of the test views.
\difix with its own Gaussian model follows its official pipeline; on \stereogs, it resumes from the trained \stereogs model as its 30k-step checkpoint, refines it to 60k steps with its release schedule, and applies its post-render step.
\seva runs at a short side of 576 pixels.

\paragraph{Pseudo poses.}
The camera $j$ in \Eqn{pose} is drawn from the $\min(3,N-1)$ cameras nearest to camera $i$.

\section{Additional Analysis}
\label{sec:supp:analysis}

\paragraph{Kernel families.}
\Fig{kernel_all} compares all twelve kernel families.
Among the one-parameter families, the first-order kernel fits best and gains the most; more parameters add at most $0.004$ to $R^2$ and $0.01$\,dB to the gain.

\begin{figure}[H]
  \centering
  \includegraphics[width=\linewidth]{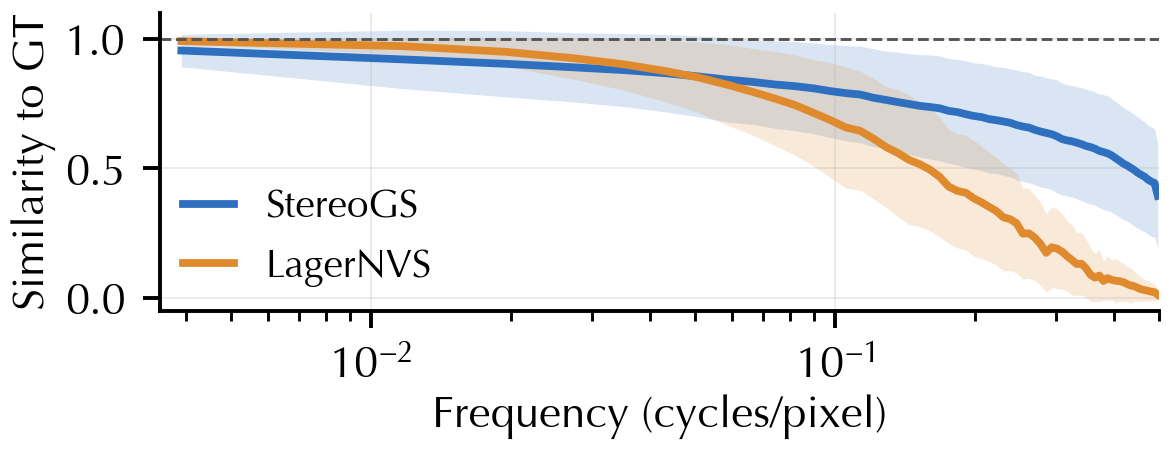}
  \caption{Similarity to GT per frequency over all 42 settings.}
  \label{fig:spectrum_all}
\end{figure}

\paragraph{Generality of the frequency pattern.}
Over all 42 settings, the error of \lager rises faster with frequency than that of the Gaussians, so the curves cross once (\Tbl{slope}, \Fig{spectrum_all}).

\begin{table}[H]
  \centering
  \footnotesize
  \setlength{\tabcolsep}{4pt}
  \begin{tabular}{lcc}
    \toprule
    Model & \begin{tabular}[b]{@{}c@{}}Error rise (dB) per\\$10\times$ frequency\end{tabular} & \begin{tabular}[b]{@{}c@{}}Crossover with\\\lager (cycles/pixel)\end{tabular} \\
    \midrule
    \stereogs & 5.2 & 0.053 \\
    \dropg & 4.1 & 0.084 \\
    \fsgs & 4.1 & 0.094 \\
    \lager & 7.4 & -- \\
    \bottomrule
  \end{tabular}
  \caption{How much the relative error rises when the frequency grows tenfold, and where it crosses that of \lager.}
  \label{tab:slope}
\end{table}

\paragraph{Sensitivity to $f_0$.}
Our method is insensitive to $f_0$ (\Tbl{sens}).
A wide range of $f_0$ keeps the gain of \teacher within $0.05$\,dB of its best, a random choice of the development set changes the gain little, and the $f_0$ of \stereogs transfers to the other two base methods.

\begin{table}[H]
  \centering
  \footnotesize
  \setlength{\tabcolsep}{4pt}
  \begin{tabular}{lccc}
    \toprule
    & \stereogs & \dropg & \fsgs \\
    \midrule
    Flat range of $f_0$ & 0.022--0.057 & 0.038--0.091 & 0.046--0.109 \\
    Median, 200 dev sets & 1.21 & 1.59 & 1.61 \\
    5th pct., 200 dev sets & 1.07 & 1.47 & 1.51 \\
    Own $f_0$ & 1.22 & 1.60 & 1.62 \\
    $f_0$ of \stereogs & 1.22 & 1.56 & 1.55 \\
    \bottomrule
  \end{tabular}
  \caption{PSNR gain (dB) of \teacher for different choices of $f_0$; in the flat range it stays within $0.05$\,dB of the best.}
  \label{tab:sens}
\end{table}

\begin{table}[H]
  \centering
  \footnotesize
  \setlength{\tabcolsep}{2.5pt}
  \begin{tabular}{lccccc}
    \toprule
    Method & LLFF-3 & LLFF-6 & LLFF-9 & Mip-12 & Mip-24 \\
    \midrule
    \rowcolor{rowlit} \multicolumn{6}{l}{\textit{Feed-forward Methods}} \\
    \anysplat & 818 & 475 & 345 & 374 & 142 \\
    \depthsplat & 420 & 243 & 163 & 129 & 127 \\
    LVSM & 7.7 & 3.0 & 1.6 & 1.0 & 0.3 \\
    \lager & 57.3 & 38.3 & 30.8 & 31.8 & 17.9 \\
    \midrule
    \rowcolor{rowlit} \multicolumn{6}{l}{\textit{Diffusion-guided Gaussians}} \\
    \stereogs + \difix & 6.1 & 6.1 & 6.1 & 7.3 & 7.3 \\
    \midrule
    \rowcolor{rowlit} \multicolumn{6}{l}{\textit{Sparse-view Gaussians with Ours}} \\
    \stereogs & 882 & 892 & 899 & 1000 & 741 \\
    \rowcolor{rowdistill} \stereogs + \distilled & 922 & 922 & 924 & 1036 & 781 \\
    \midrule
    \dropg & 1379 & 1100 & 701 & 1046 & 524 \\
    \rowcolor{rowdistill} \dropg + \distilled & 1377 & 1094 & 702 & 1048 & 527 \\
    \midrule
    \fsgs & 1024 & 1092 & 790 & 962 & 532 \\
    \rowcolor{rowdistill} \fsgs + \distilled & 976 & 1062 & 778 & 949 & 530 \\
    \bottomrule
  \end{tabular}
  \caption{Rendering speed in frames per second at the evaluation resolution on one RTX 5090, averaged over all settings of each group (all nine Mip-NeRF 360 scenes).}
  \label{tab:speed}
\end{table}

\begin{figure}[H]
  \centering
  \includegraphics[width=0.9\linewidth]{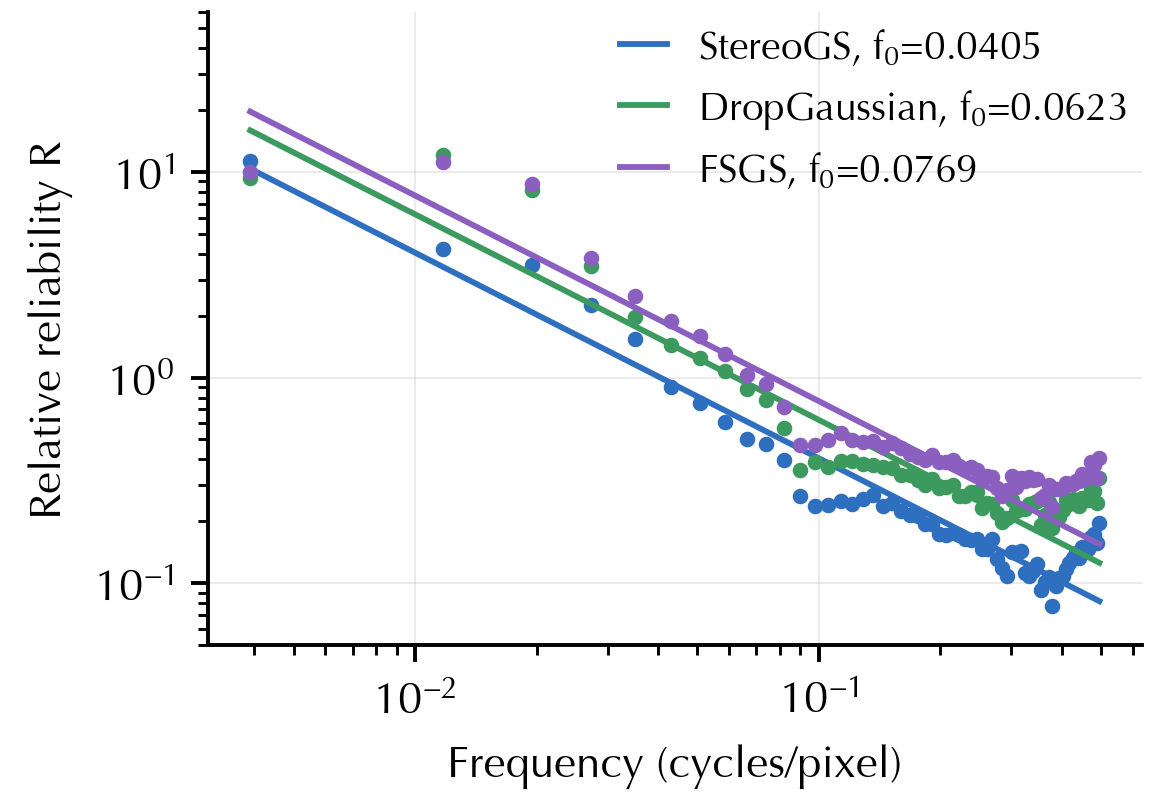}
  \caption{Relative reliability $R=w^*/(1-w^*)$ of the prior for three bases on the first development set, with $f_0/f$ lines. Below $0.2$ cycles per pixel the slope is $-1.1$ to $-1.2$.}
  \label{fig:reliability}
\end{figure}

\paragraph{The $1/f$ law.}
The first-order kernel has a simple meaning.
We call $R=w^*/(1-w^*)$ the relative reliability of the prior: it says how many times more the optimal fusion trusts the prior than the Gaussians at a frequency.
With the kernel of \Eqn{kernel} as the weight, $K/(1-K)=f_0/f$, so the kernel assumes that this relative reliability falls as $1/f$ and equals one at the crossover $f_0$.
The measured $R$ indeed follows $1/f$ for all three base methods (\Fig{reliability}), which is why a first-order kernel fits the optimal weight so well.

\paragraph{Detailed speed numbers.}
\Tbl{speed} lists the rendering speed of every method for each dataset and view count, which \Fig{teaser} and \Tbl{fps} summarize.
Distillation keeps every base method within $5\%$ of its own speed in every group, while all methods that run a network per frame stay below 60\,FPS.

\section{Choice of the Prior: Details}
\label{sec:supp:prior}

We run the study of \Sect{prior} on the 24 LLFF settings, with \stereogs as the base and everything else fixed, and fit $f_0$ for each prior on the LLFF settings of the two development sets.
The first-order kernel is the best one-parameter family for every prior, and its lead grows  for weaker prior (\Tbl{priorfam}).

\begin{table}[H]
  \centering
  \footnotesize
  \setlength{\tabcolsep}{4pt}
  \begin{tabular}{lcccc}
    \toprule
    Prior & Ideal low-pass & Gaussian & Exponential & First-order \\
    \midrule
    \lager & 0.812 & 0.937 & \underline{0.987} & \textbf{0.992} \\
    \seva & 0.584 & 0.870 & \underline{0.954} & \textbf{0.984} \\
    \lvsm & 0.074 & 0.488 & \underline{0.608} & \textbf{0.807} \\
    \bottomrule
  \end{tabular}
  \caption{Fit quality $R^2$ of the one-parameter kernel families to the optimal weight of each prior, on the first LLFF development set.}
  \label{tab:priorfam}
\end{table}

\section{Discussion}
\label{sec:supp:discussion}

\paragraph{Deployment on resource-constrained devices.}
The distilled model is a standard 3DGS model: distillation changes the parameters of the base model and adds no Gaussians, so it renders fast on a server GPU.
Deploying it on mobile or other resource-constrained devices may need system optimizations or dedicated accelerators~\cite{feng2025lumina,feng2024potamoi,feng2025streamgrid}, or streaming from a server~\cite{wu2025immersive}.
For example, we can track the gaze of the user~\cite{zhao2026gazeflow} and lower the rendering cost and power with foveated rendering~\cite{lin2025metasapiens,lin2026metasapiensv2, lin2025powergs}.
We can also tone-map the rendered images to lower the display power on power-constrained devices such as AR headsets~\cite{lin2026lowpowar}, and condition the Gaussians on such downstream processing~\cite{lin2026controlgs}.

\paragraph{Video diffusion priors.}
\seva, which we test as a prior in \Sect{prior}, is a multi-view diffusion model, but a video diffusion model can also serve as the prior.
For example, \mode{ControlHair}~\cite{lin2026controlhair} drives a video diffusion model with control signals to render bullet-time novel views.
We can also leverage attention modulation to keep the subject consistent across views~\cite{zhao2026referenceforcing}.

\paragraph{Perceptual evaluation.}
We report PSNR, SSIM and LPIPS, but perceived image quality has many more dimensions than these metrics capture~\cite{zhao2026observer}.
A perceptual study of the fused and distilled images is future work.

\end{document}